# Rank Selection of CP-decomposed Convolutional Layers with Variational Bayesian Matrix Factorization

Marcella Astrid[1], Seung-Ik Lee[1,2], Beom-Su Seo[2]

[1] University of Science and Technology, Daejeon, South Korea
[2] Electronics and Telecommunications Research Institute, Daejeon, South Korea
marcella.astrid@ust.ac.kr, the_silee@etri.re.kr, bsseo@etri.re.kr

*Abstract*— **Convolutional Neural Networks (CNNs) is one of successful method in many areas such as image classification tasks. However, the amount of memory and computational cost needed for CNNs inference obstructs them to run efficiently in mobile devices because of memory and computational ability limitation. One of the method to compress CNNs is compressing the layers iteratively, i.e. by layer-by-layer compression and fine-tuning, with CP-decomposition in convolutional layers. To compress with CP-decomposition, rank selection is important. In the previous approach rank selection that is based on sensitivity of each layer, the average rank of the network was still arbitrarily selected. Additionally, the rank of all layers were decided before whole process of iterative compression, while the rank of a layer can be changed after fine-tuning. Therefore, this paper proposes selecting rank of each layer using Variational Bayesian Matrix Factorization (VBMF) which is more systematic than arbitrary approach. Furthermore, to consider the change of each layer's rank after fine-tuning of previous iteration, the method is applied just before compressing the target layer, i.e. after fine-tuning of the previous iteration. The results show better accuracy while also having more compression rate in AlexNet's convolutional layers compression.**

## I. INTRODUCTION

Convolutional Neural Networks (CNNs) for computer vision tasks have shown notable results. CNNs capability in image classification was introduced by AlexNet [1] that won 2012 ImageNet Large-Scale Visual Recognition Challenge (ILSVRC). Since then, various CNN-based image classification appears, such as VGG [2] and GoogleNet [3] that achieve around 90% of top-5 classification accuracy. Besides image classification, CNNs have also shown notable results in other tasks, such as object detection [4]-[6], age estimation [7], and object tracking [8], [9].

With such capability, there is an increasing demand to deploy CNNs on smart devices such as mobile phone. However, CNNs will need high amounts of memory and computational resources, which are hindrance of the deployment because of resource limitation in smart devices.

The solution of the problem is model compression by decreasing its size and computational size. Various approaches have been proposed, such as pruning [10], quantization [11]-[12], matrix factorization [13]-[14], and tensor decomposition [15]-[18]. We select tensor decomposition-based method because it is straight-forward to the convolutional layer structure. Some approaches have been proposed based on tensor decomposition, such as Tucker decomposition [15] and CP-decomposition [16]-[18]. CP-decomposition seems to compress much compare to Tucker decomposition because it does not have core tensor in its decomposed form.

To solve CP-decomposition instability [17] problem in all layers compression, Astrid and Lee [18] iteratively decomposed the layer. Shown better result compare to Tucker decomposition [15] in AlexNet compression, the approach has deficiency in rank selection while it is important as it affects the compression rate as well as the accuracy. Too high rank will not maximize the compression, while too low may make the accuracy recovery difficult or impossible.

Two problems in the previous work's rank selection method, called sensitivity approach, are the arbitrariness of average rank selection and rank selection is done before the whole process begins. The first problem makes the implementation is based on trial and error. The second problem exists because after each iteration, i.e. fine-tuning whole network after one layer compression, the value of weights are changed. The changing value may affect the rank of the next layers. Therefore, we need to find rank of a layer after fine-tuning of previous iteration.

To solve the problems, this paper proposes Variational Bayesian Matrix Factorization (VBMF) [19] to estimate rank of a tensor. The global analytic VBMF is a propitious method that can estimate noise of a matrix before recover the matrix rank. Kim et al. [15] have applied this method for finding Tucker decomposition rank. However, CP-decomposition is different as each tensor only needs one rank while Tucker not. The method is applied on AlexNet convolutional layer as only convolutional layers are decomposed by CP-decomposition.

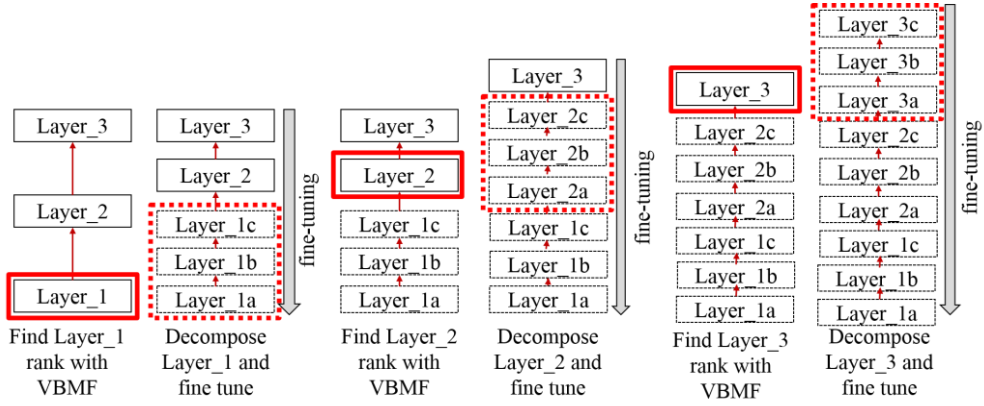

**Figure 1.** Iterative compression and fine-tuning.

## II. METHODOLOGY

Our approach using iterative compression to compress all convolutional layer which is modification of previous work [18]. Rank calculation and compression of a convolutional layer, followed by whole layer fine-tuning are done in each iteration. Therefore, this section will be separated into three parts: iterative compression and fine-tuning, compression of convolutional layer, and rank selection.

### A. Iterative Compression and Fine-Tuning

Astrid and Lee [18] have shown empirically that the iterative compression and fine-tuning can solve CP instability problem to compress all layers. However, by using sensitivity method to decide the rank, the rank change after fine-tuning is not considered. Therefore, in our work, as seen in Figure 1, rank selection of a layer is done every iteration before compression of the corresponding layer and fine-tuning of whole network. Since the rank selection is done independently for each layer, in contrast with sensitivity approach that based on ratio of each layer sensitivity, VBMF is used to estimate the rank.

### B. CP-Decomposition for Convolutional Layer Compression

Tensor is generalization of matrix. As seen in Figure 2 left, a convolutional layer can be seen as 3-way tensor, i.e. 3-D matrix, of size $(D \times D) \times S \times T$ where $D$ is spatial dimension of filter, $S$ is input channel, and $T$ is output channel or the number of filters. The $D \times D$ spatial dimension is combined as one dimension.

After CP-decomposition, as seen in Figure 2 right, a layer of convolutional layer becomes three layers of convolutional layers of size $1 \times 1 \times S \times R$, $D \times D \times 1 \times R$, and $1 \times 1 \times R \times T$ where $R$ is the rank of CP-decomposition that we need to estimate.

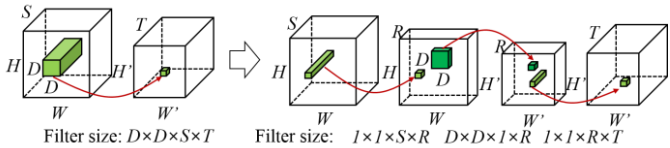

**Figure 2.** A convolutional layer is decomposed by CP-decomposition

For more detailed information about the CP-decomposition for convolutional layer compression, we refer the reader to the previous work [18].

### C. Tensor Rank Estimation with VBMF

Nakajima et al. [19] shows global analytic solution of VBMF. It automatically denoises the matrix under a low-rank assumption, therefore, it can find rank of denoised low-rank matrix. Since VBMF can find matrix rank instead of tensor, the weight tensors have to be converted into matrices, which process is called matricization. Matricization of a 3-way tensor can be done in three types, such in Figure 3, as there are three ways to slice the tensor. Each cases looks for rank in $T$ dimension (Figure 3 (a)), $S$ dimension (Figure 3 (b)), and $D^2$ (Figure 3 (c)) dimension respectively.

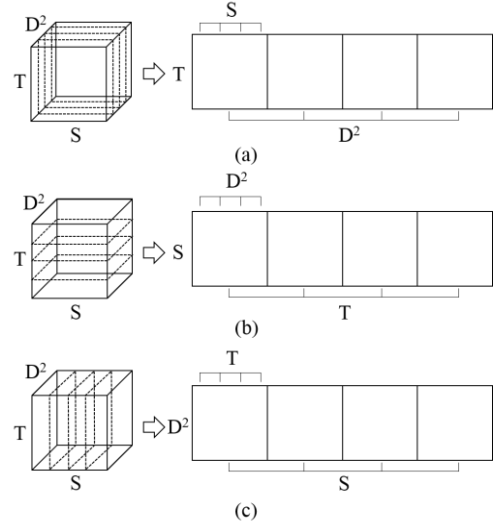

**Figure 3.** Three types of 3-way tensor matricization of a 3-way tensor. (a) Type 1 of size $T \times SD^2$. (b) Type 2 of size $S \times TD^2$. (c) Type 3 of size $D^2 \times TS$.

After calculating rank for each matrix type, we selected the maximum rank of all the three as we only need one rank for the tensor rank. Maximum rank is selected to make sure the restorability of the accuracy. If selecting lower than the maximum rank, there is higher probability that one or more dimension are compressed too much which can lead to accuracy loss or not-recoverable accuracy.

## III. EXPERIMENTS

We test our approach on one of representative CNN, AlexNet [1], and using Caffe framework [20]. Before explaining the experiments result, we briefly introduce AlexNet.

### A. AlexNet Overview

AlexNet is the winner of ILSVRC 2012 that capable to classify 1,000 classes from ImageNet 2012 dataset [21]. The top-1 and top-5 accuracy from 50,000 validation images are 56.83% and 79.95%, respectively. It has five convolutional layers followed by three fully connected layers. The second, fourth, and fifth convolutional layers are branched into two "streams" (group = 2) to split the training because the training process was computationally too expensive. In this work, the first convolution is considered as 2-way tensor of size $T \times SD^2$ because the S dimension is small ($S=3$) so it is combined with the spatial dimension.

### B. Rank Selection

Table 1 shows the calculation of each AlexNet layer with publicly available VBMF [22]. In the layer 2, 4, and 5, as they have group equals to 2, the ranks are calculated for each group tensor.

**TABLE 1.** ESTIMATED RANK OF EACH CONVOLUTIONAL LAYER IN ALEXNET FROM 3 MATRICIZATION TYPES.

| Layer | Group | Type 1 ($T \times SD^2$) | Type 2 ($S \times TD^2$) | Type 3 ($D^2 \times TS$) | Max |
|---|---|---|---|---|---|
| Conv1 | - | 53 | - | - | 53 |
| Conv2 | - | - | - | - | 83×2 |
| | 1 | 83 | 30 | 16 | - |
| | 2 | 78 | 25 | 17 | - |
| Conv3 | - | 138 | 103 | 6 | 138 |
| Conv4 | - | - | - | - | 119×2 |
| | 1 | 37 | 35 | 7 | |
| | 2 | 34 | 119 | 7 | |
| Conv5 | - | | | | 109×2 |
| | 1 | 32 | 40 | 6 | |
| | 2 | 28 | 109 | 6 | |

### C. Convolutional Layer Compression

Table 2 shows the convolutional layers compression result in AlexNet. To make fair comparison, we also compress only convolutional layers, instead of whole layers, using sensitivity approach rank selection from the previous work [18]. As convolutional layers have shared weights, the number of weights is not large, therefore the compression too. The compression focuses more on the computational cost compression.

This work achieves higher accuracy compare to previous sensitivity rank selection approach, even though it has more compression, theoretically and with Caffe CPU time. Even, the accuracy increases around 1% compare to the original network.

## IV. CONCLUSION

This work proposes VBMF to select rank of CP-decomposed convolutional layers. Rank selection is done in every compression and fine-tuning iteration to consider the weights change after fine-tuning that may affect the rank change. The result surpasses the previous sensitivity selection by compressing more with better accuracy. Specifically, our method achieves top-1 and top-5 accuracy of 55.71% and 81.03%, respectively, which are higher than the original network and previous approach, while having higher compression rate. For future study, compressing fully-connected layer systematically is interesting to be explored.

## ACKNOWLEDGEMENT

This work was supported by Institute for Information & communications Technology Promotion (IITP) grant funded by the Korea government (MSIT) (No.2017-0-00067, Development of ICT Core Technologies for Safe Unmanned Vehicles).

**TABLE 2.** CONVOLUTIONAL LAYERS COMPRESSION RESULTS

| | Top-1 Acc | Top-5 Acc | Weights | Theory Cost | CPU Time (ms) |
|---|---|---|---|---|---|
| Original | 56.83 | 79.95 | 61.0M | 724M | 167.71 |
| CP-sensitivity [18] (Ave rank: Conv 150) | 55.71 (-1.12) | 79.30 (-0.65) | 58.9M (×1.03) | 255M (×2.84) | 125.63 (×1.34) |
| This work | 58.35 (+1.52) | 81.03 (+1.08) | 58.9M (×1.03) | 238M (×3.05) | 121.41 (×1.38) |

**Marcella Astrid** received the BS in computer engineering from Multimedia Nusantara University, Tangerang, Indonesia, in 2015, and the MEng in computer software from University of Science and Technology (UST), Daejeon, South Korea, in 2017, and in the same university, is currently working toward PhD degree in computer science. Her recent interests include deep learning and computer vision.

**Seung-Ik Lee** received his MS and PhD in computer science from Yonsei University, Seoul, South Korea, in 1997 and 2001, respectively. He is currently working for ETRI (Electronics and Telecommunications Research Institute), South Korea. His research interests include machine learning, deep learning, and reinforcement learning.

**Beom-Su Seo** received his MS in computer science and statistics from University of Seoul, South Korea, in 1998. He is currently working for ETRI (Electronics and Telecommunications Research Institute), South Korea as a project leader on the environment perception for unmanned vehicles and the safety verification and validation for service robots.